\documentclass[letterpaper, 10 pt, conference]{ieeeconf}  

\IEEEoverridecommandlockouts                              

\usepackage{cite}
\usepackage{amsmath,amssymb,amsfonts}

\usepackage{graphicx}
\usepackage{textcomp}
\usepackage{xcolor}
\usepackage{hyperref}
\usepackage{multirow}
\usepackage{hhline}
\usepackage{subcaption}
\usepackage{algorithm}
\usepackage{algorithmic}
\usepackage{array}
\usepackage{stfloats}
\usepackage{url}
\usepackage{enumerate}
\usepackage{float}
\usepackage[normalem]{ulem}
\usepackage{yhmath}
\usepackage[export]{adjustbox}
\usepackage{diagbox}
\usepackage{kotex}
\usepackage{verbatim}
\usepackage[margin=0.75in]{geometry}
\usepackage{gensymb}
\renewcommand{\baselinestretch}{0.91}
\usepackage{amsmath}
\newcommand\norm[1]{\left\lVert#1\right\rVert}

\usepackage{caption}
\title{
\bf
Towards Defensive Autonomous Driving: \\ 
Collecting and Probing Driving Demonstrations of Mixed Qualities
}

\author{Jeongwoo Oh$^{1*}$, Gunmin Lee$^{1*}$, Jeongeun Park$^2$, Wooseok Oh$^1$, Jaeseok Heo$^1$, Hojun Chung$^3$, \\
Do Hyung Kim$^4$, Byungkyu Park$^4$, Chang-Gun Lee$^4$, Sungjoon Choi$^2$, and Songhwai Oh$^1$
\thanks{*Equal contribution}
\thanks{$^{1}$J. Oh, G. Lee, W. Oh, J. Heo, and S. Oh are with the Department of Electrical and Computer Engineering and ASRI, Seoul National University, Seoul 08826, Korea (e-mail:  jeongwoo.oh@rllab.snu.ac.kr, gunmin.lee@rllab.snu.ac.kr, wooseok.oh@rllab.snu.ac.kr, jaeseok.heo@rllab.snu.ac.kr, songhwai@snu.ac.kr).}%
\thanks{$^{2}$J. Park, and S. Choi are with the Department of Artificial Intelligence, Korea University, Seoul 02841, Korea (e-mail:baro0906@korea.ac.kr, sungjoon-choi@korea.ac.kr).}%
\thanks{$^{3}$H. Chung is with the Department of Mechanical Engineering, Seoul National University, Seoul 08826, Korea (e-mail:cssens@snu.ac.kr).}
\thanks{$^{4}$D. Kim, B. Park, and C. Lee are with the Department of Computer Science and Engineering, Seoul National University, Seoul 08826, Korea (email:totheparadise119@gmail.com, earthquake16@gmail.com, cglee@snu.ac.kr).}
}

\begin{document}

\maketitle
\thispagestyle{empty}
\pagestyle{empty}

\begin{abstract}
Designing or learning an autonomous driving policy is undoubtedly a challenging task as the policy has to maintain its safety in all corner cases. In order to secure safety in autonomous driving, the ability to detect hazardous situations, which can be seen as an out-of-distribution (OOD) detection problem, becomes crucial. However, most conventional datasets only provide expert driving demonstrations, although some non-expert or uncommon driving behavior data are needed to implement a safety guaranteed autonomous driving platform. To this end, we present a novel dataset called the \textit{R3 Driving Dataset}, composed of driving data with different qualities. The dataset categorizes abnormal driving behaviors into eight categories and 369 different detailed situations. The situations include dangerous lane changes and near-collision situations. To further enlighten how these abnormal driving behaviors can be detected, we utilize different uncertainty estimation and anomaly detection methods to the proposed dataset. From the results of the proposed experiment, it can be inferred that by using both uncertainty estimation and anomaly detection, most of the abnormal cases in the proposed dataset can be discriminated. 
The dataset of this paper can be downloaded from https://rllab-snu.github.io/projects/R3-Driving-Dataset/doc.html.

\end{abstract}

\section{INTRODUCTION}
\label{intro}

It has been a challenging task to design a controller for autonomous driving. Conventional approaches to design a controller are based on rule-based methods \cite{rulebased1, rulebased2, rulebased3, rulebased4}, reinforcement learning (RL) \cite{RLbased1, RLbased2, RLbased3, RLbased4}, and imitation learning (IL) \cite{ILbased1, ILbased2, ILbased3, ILbased4}. However, rule-based methods are not proficient at covering all exception cases due to the diversity of exceptions in the open road. In the case of RL-based methods, a stable and elaborate reward function is required to learn a controller. For an IL-based method, the algorithm exploits and tries to learn to behave similarly to the provided data. Since IL-based controller follows the provided driving data, if it is provided with more good driving data, its performance can be improved further. In addition, an IL method has an advantage over an RL method as it does not require a reward function. These characteristics allow an IL-based method to show high performance. Therefore, in this paper, we focus on the IL methods.

Designing a controller via IL requires driving data. As autonomous driving is a safety-critical problem, both expert and abnormal data are needed to design the controller to prepare for abnormal driving cases. Without abnormal driving demonstrations, the autonomous vehicle may become vulnerable to abnormal and unseen situations. However, to the best of our knowledge, the existing datasets do not cover any extreme cases, covering only the expert and safe situations. To compensate for this issue, we propose a novel dataset called the \textit{R3 Driving Dataset}, which covers both safe driving and abnormal driving cases. The abnormal driving cases include dangerous lane changes, occupying more than one lane, and near-collision situations in crossroads and straight roads.

\begin{figure}[t]
  \centering
  \includegraphics[width=0.8\linewidth]{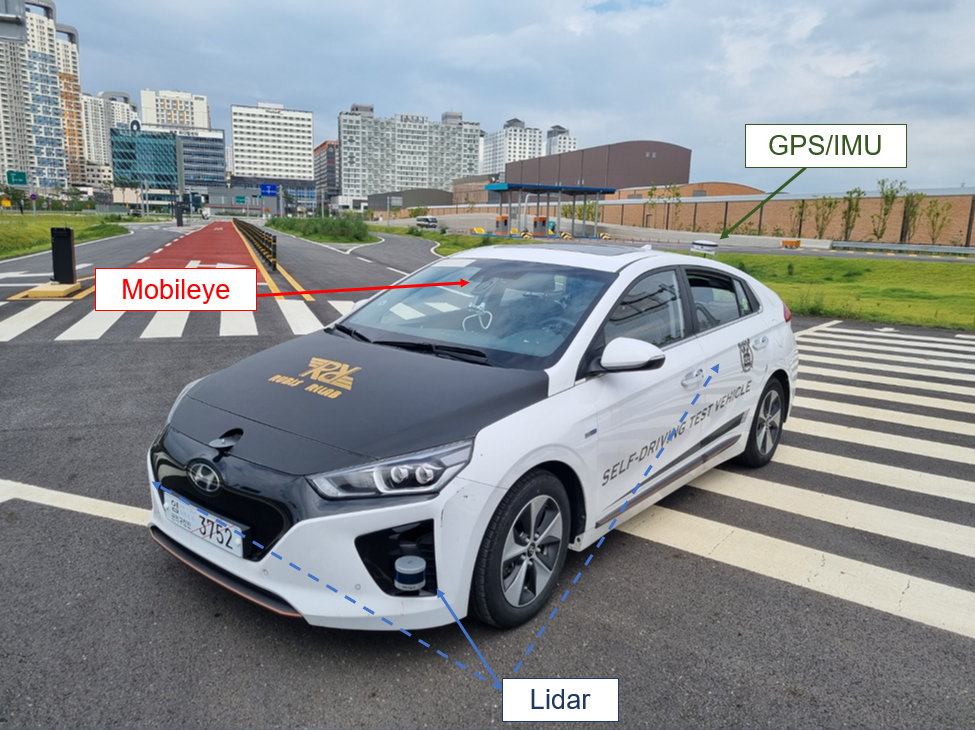} 
\caption{$\mathbf{Data~Collection~Platform.}$ The proposed Hyundai Ioniq platform is equipped with three Velodyne LiDARs (marked in blue), a Mobileye sensor (marked in red), and a combined GPS/IMU inertial navigation system (marked in green). The dotted lines indicate the occluded sensors.}
\label{platformsketch}
\end{figure}

As we want the vehicles surrounding the ego vehicle to act similar as they normally do, the ego vehicle needs to have similar appearance to normal vehicles. However, due to this assumption, we excluded the image data, as the camera needs to be installed inside the vehicle. If the camera is installed inside the vehicle, the image input has low resolution and does not cover the whole forward view, causing the quality of image to deteriorate. A photo of the proposed data collection platform is shown in Figure \ref{platformsketch}.


As the expert demonstrations are sampled state-action pairs from the expert driving policy, the expert demonstrations can be seen as data points sampled from an expert driving distribution. On the other hand, the abnormal demonstrations are sampled from an abnormal driving policy, a distinct policy to the expert driving policy. In terms of distribution, expert and abnormal demonstrations are in an in-distribution (ID) and out-of-distribution (OOD) relationship, respectively. Therefore, when discriminating expert and abnormal demonstrations, the process can be seen as anomaly detection.

We have safely collected expert demonstrations in three different locations, a highway, urban road, and Future Mobility Technology Center (FMTC) in Siheung-si, Korea. In case of abnormal demonstrations, the data are collected only in FMTC due to the safety concerns. The abnormal demonstrations are collected based on the driving accident cases listed in \cite{cadas}. The vehicle trajectories are set beforehand for safety during the data collection, and the control tower
has been alert throughout the experiment. The number of expert demonstrations collected is 42,400, and the number of abnormal demonstrations collected is 20,355. The exact figures are shown in Table \ref{tab:expert} and \ref{tab:negative}.

The proposed dataset is evaluated using four different evaluation metrics for anomaly detection. The four metrics are the epistemic uncertainty, aleatoric uncertainty, pi-entropy, and reconstruction loss. The epistemic uncertainty, aleatoric uncertainty, and pi-entropy are calculated using the mixture density network (MDN) \cite{MDN}. The reconstruction loss is calculated by using the variational autoencoder (VAE) \cite{VAEbottleneck}. 
The results show that the epistemic uncertainty best detect failing lane-keeping and unstable driving situations, which are closely related to the decision of the ego driver. On the other hand, the reconstruction loss best detects near-collision, dangerous overtaking, and dangerous lane-changing situations, which are mainly influenced by the surrounding vehicles.

\section{related work}

Several approaches to provide an autonomous driving dataset \cite{KITTI, nuscene, Waymo, A2D2, stanf, Llamas} have been made. Geiger et al. \cite{KITTI} provide the researchers with the most commonly used dataset. The dataset includes data collected via a camera, LiDAR, and GPS and has annotations of bounding boxes, semantic labels, and lane marking. Caesar et al. \cite{nuscene} provide the same categories of data to \cite{KITTI}, but have behavioral label annotations. Unlike \cite{KITTI} and \cite{nuscene}, Geyer et al. \cite{A2D2} provide additional data of other vehicles. Sun et al. \cite{Waymo} provide the researchers with all the information in \cite{KITTI, nuscene, A2D2}. Behrendt et al. \cite{Llamas} provide only the LiDAR data and annotations for lane markings. However, the above datasets only provide expert demonstrations, while this paper provides a new dataset with abnormal driving situations. 

A number of methods \cite{VAEbottleneck, OODbase, ODIN, TCC, uncanomaly} are applied to discriminate OOD samples from in-distribution samples. An et al. \cite{VAEbottleneck} proposed a method using the reconstruction loss of VAE as a metric for OOD detection. Hendrycks et al. \cite{OODbase} proposed an evaluation method and baselines for the OOD detection. The baselines are called the area under the receiver operating characteristic curve (AUROC) and area under the precision-recall curve (AUPR). Liang et al. \cite{ODIN} proposed a method using temperature scaling and an input preprocessing method to discriminate unknown samples without modifying the existing structure of the neural network model. Lee et al. \cite{TCC} proposed to modify the loss function by adding the GAN loss and confidence loss terms to the cross-entropy loss in order to discriminate an OOD sample during the training process. Islam et al. \cite{uncanomaly} proposed a method to discriminate OOD samples using uncertainty values.  

Yannis et al. \cite{cadas} provided the statistics about vehicle accident cases in Europe. The author categorized the traffic accident cases into 44 different categories. The vehicle accident types include pedestrians and vehicle collisions in both crossroads and straight roads, collisions between vehicles on straight roads and crossroads, and single-vehicle accidents in straight, curved, and crossroads. The proposed dataset in this paper follows the categorizations in \cite{cadas}. 

\section{Background}

\subsection{Data Collection}
\subsubsection{Object Detection}
\label{sec:ob_detect}
For object detection, LiDAR sensors are used, which obtain point cloud information of surroundings. Our object detection algorithm consists of two clusterings and one gradient descent method. The proposed detection algorithm is based on \cite{clustering}. 

Before the clustering, the obtained point cloud is preprocessed to exclude the ground data, which is redundant for the clustering algorithm. The first clustering algorithm is the Euclidean clustering, which is used to cluster points in the point cloud. To speed up the clustering algorithm, a voxel tree data structure is used. After that, a convex hull is built for each cluster. The second Euclidean clustering is performed among the convex hulls to solve multiple clusters from the same object in the blind spot of LiDARs or when the objects are far from the LiDAR. Similar to the first Euclidean clustering, the second clustering algorithm considers convex hulls within a certain distance to be the same cluster, and the distances between any two convex hulls are calculated using the Gilbert–Johnson–Keerthi distance algorithm \cite{GJK} to reduce time complexity. After grouping one cluster per object, the object parameters are optimized by the gradient descent method using the following loss function: 
\begin{equation}
\label{eq:object_detect}
\begin{aligned}
L & = \alpha_\text{w} (w-w_\text{g})^2+\alpha_\text{l} (l-l_\text{g})^2 \\
& +\frac{\alpha_\text{p}}{N} \sum_{i=1}^N d_{\text{box}}(p_\text{i})^2 \\
& +\alpha_{\text{cvh}}(d_{\text{cvh}}(O_{\text{ego}}) - {d_{\text{box}}(O_{\text{ego}}}))^2, \\
\end{aligned}
\end{equation}
where $w$ and $l$ are the width and length of the bounding box, $w_\text{g}$ and $l_\text{g}$ are the desired width and length of a vehicle, $d_{\text{cvh}}(p)$ and $d_{\text{box}}(p)$ are the distance to the convex hull and the bounding box from point $p$, $p_i$ is the $i$th point of point cloud, which is a convex hull, $N$ is the number of valid vertices on the convex hull, $O_{\text{ego}}$ is the origin of the ego vehicle, and $\alpha_\text{p}$, $\alpha_\text{w}$, $\alpha_{\text{cvh}}$ and $\alpha_\text{l}$ are hyperparameters. We treat the point on the convex hull as valid only when it is not occluded from the ego vehicle. The overall complexity is O($N_{all}$ + $N_{r}^2$), where $N_{all}$ is the number of points in one frame, and $N_{r}$ is the number of redundant points after removing the ground.

\subsubsection{Object Tracking}
With only one time frame of the point cloud, the dynamic information of vehicles cannot be estimated. However, dynamic information such as linear velocity, linear acceleration, and angular velocity of surrounding vehicles in the driving environment is essential for identifying dangerous situations. Therefore, an object tracking algorithm using the extended Kalman filter (EKF) \cite{EKF} is implemented to estimate dynamic information.

The state information of an object used to calculate EKF consist of nine dimensions. The information includes the location and position of the objects (three dimensions), calculated from the Cartesian coordinate where the origin is the ego vehicle. The lineal and lateral directions of the ego vehicle are the x-axis and the y-axis, respectively. The information also includes the linear velocity, linear acceleration, and angular velocity of the objects and the linear velocity, linear acceleration, and angular velocity of the ego vehicle.

The observed object list is given at each timestep in the object tracking algorithm and the current object tracking list is updated. First, the next frame states of the objects in the current object tracking list are predicted. Next, we apply the cost function proposed in \cite{object_tracking} in order to decide the matching order of object pairs between the current and observed object lists. 

After the cost calculation, similar to \cite{object_tracking}, the objects are matched greedily until a certain threshold for the calculated cost is met. The objects are matched from the lowest to highest cost. The objects that are matched in the current object list go through the update step of EKF to estimate the next state. For the objects that are not matched in the current object list, the state information from the prediction is used to estimate the next state. If the object is not in the current object list for five consecutive frames, it is removed from the current object list. The objects that are not matched in the observed object list are thought of as new objects and inserted into the current object list.

\subsection{Prediction Uncertainty in a Mixture Density Network}
\label{sec:mdn}
One of the most popular methods to estimate the uncertainty is the mixture density network (MDN) \cite{MDN}. A MDN is composed of multiple Gaussian models, whose parameters are outputs of a neural network. The MDN with the output $\mathrm{y}$ and input $\mathrm{x}$ can be written as:
\begin{equation}
    p(\mathrm{y}|\mathrm{x}) = \sum_{j=1}^{\mathrm{K}} \pi_j(\mathrm{x})\mathcal{N}(\mathrm{y};\mu_j(\mathrm{x}), \Sigma_j(\mathrm{x})),
\label{eq:MDN}
\end{equation}
where $K$ is the number of mixtures, $\pi_j(\mathrm{x}), \mu_j(\mathrm{x}), \Sigma_j(\mathrm{x})$ are the $j$th mixture weight function, mean function, and variance function of a Gaussian mixture model (GMM).

The total uncertainty of an MDN is the variance of the MDN network. The variance can be calculated by the sum of aleatoric and epistemic uncertainties \cite{ualfd}.
The total uncertainty for input x can be calculated as:
\begin{equation}
\label{uncertaintyeq}
\begin{split}
    V(\mathrm{y}|\mathrm{x}) = & \sum_{j=1}^{\mathrm{K}}\pi_j(\mathrm{x})\Sigma_j(\mathrm{x}) + \\ & \sum_{j=1}^{\mathrm{K}}\pi_j(\mathrm{x})\norm{\mu_j(\mathrm{x})-\sum_{k=1}^{\mathrm{K}}\pi_k(\mathrm{x})\mu_k(\mathrm{x})}^2,
\end{split}
\end{equation}
where $\pi_k$ and $\mu_k$ is the $k$th mixture weight function and mean function of the GMM. The other notations are same as (\ref{eq:MDN}).

Three types of different uncertainties are used to evaluate the dataset. The first uncertainty is the epistemic uncertainty. The epistemic uncertainty is the second term of the right-hand side of (\ref{uncertaintyeq}). This uncertainty shows the difference between the predicted value of the $j$th Gaussian and the other Gaussians. Therefore, this can be seen as calculating how each predicted value is different from others. 

The second uncertainty is the aleatoric uncertainty, which is the first term of the right-hand side of the (\ref{uncertaintyeq}). This uncertainty is the weighted sum of the model variance, which can be seen as capturing the noise in the given dataset.

The third uncertainty is the pi-entropy, which denotes the entropy of mixture weights. This term is similar to epistemic uncertainty, but differs from the sense that it does not consider any disagreement among $\mu$. The equation for calculating the pi-entropy is shown as:

\begin{equation}
    \text{pi-entropy} = - \sum_{j=1}^K \pi_{j}(\mathrm{x})
    \log\left(\pi_j(\mathrm{x})\right),
\end{equation}

\subsection{Using Reconstruction Loss as a Measurement of Uncertainty}
A variational autoencoder (VAE) \cite{VAE} is composed of an encoder model and a decoder model. An encoder model can be seen as a neural network that reduces the dimension of input data. During this dimension reduction process, the encoder extracts features that are crucial to the given data. A decoder model can be seen as a neural network that reconstructs the encoded features back to the original dimension. The encoder-decoder model is trained at once by comparing the original data and the reconstructed data. The model learns to reconstruct the original data with the least amount of data loss.

An et al. \cite{VAEbottleneck} proposed that during the encoding process, the encoder selects essential features, as the loss of data is inevitable during the dimension deduction process. However, when the encoder is introduced to OOD data, the encoder selects different features to extract. When these features are reconstructed, the reconstruction result is distinct from the input data, causing the reconstruction loss to elevate. Therefore, the reconstruction loss can be used as a method for anomaly detection. 

\begin{figure}[t]
  \centering
  \includegraphics[width=0.8\linewidth]{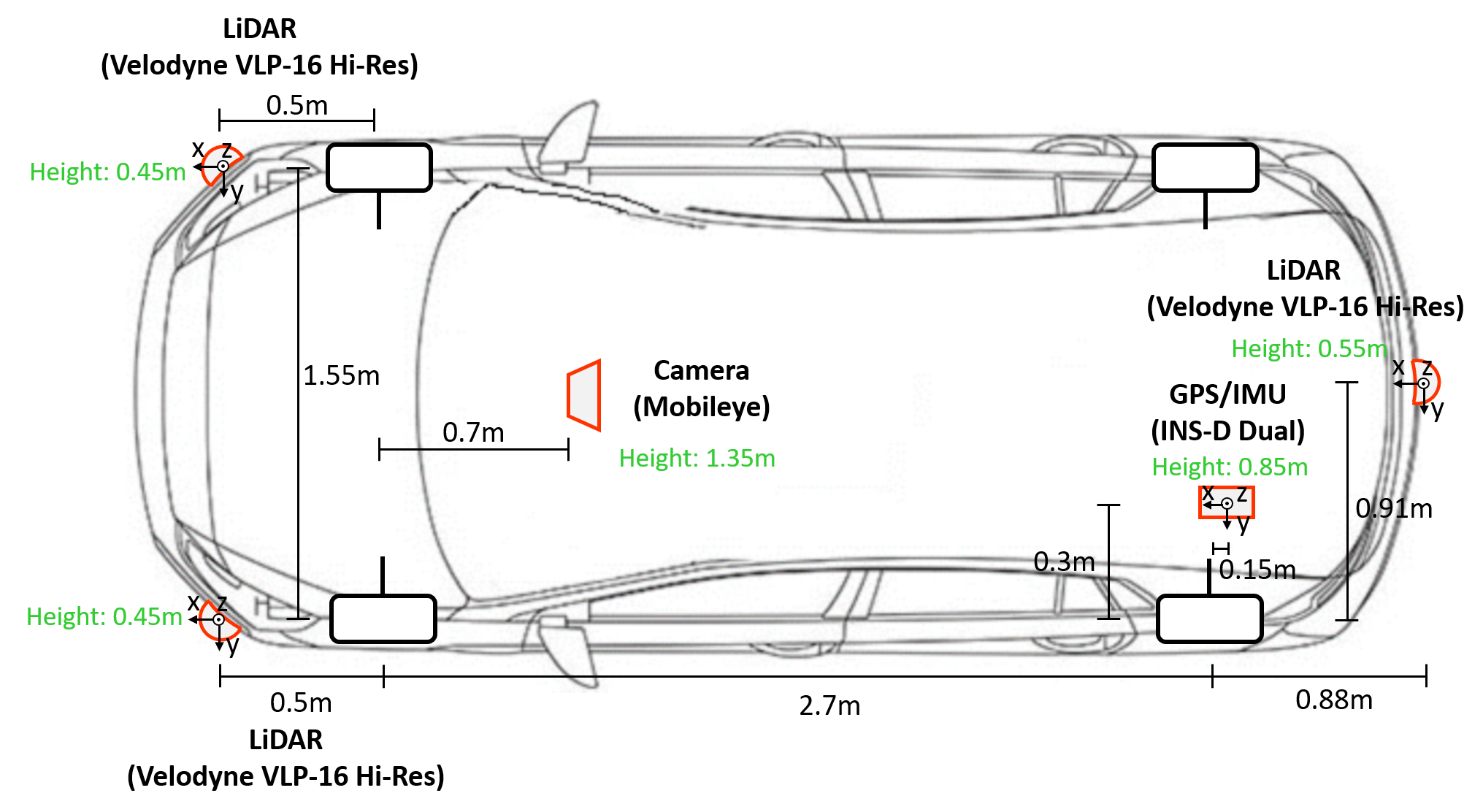}
\caption{$\mathbf{Sensor~Setup.}$ The illustration of the dimensions and the installation positions of the data collection platform.}
\label{fig:sensor}
\end{figure}

\section{experiment setup}

\subsection{Hardware Setting}
As shown in Figure \ref{fig:sensor}, various sensors are deployed on the tested vehicle (Hyundai Ioniq EV). Three LiDAR modules (Velodyne VLP-16 Hi-Res), two in the front and one in the back, are dedicated to collect surrounding environment information. These modules are used to collect the point cloud, which is exploited to gain object information around the ego vehicle. The GNSS sensor (Inertial Lab INS D) is placed on the vehicle’s base link (in the middle of the trunk). This sensor receives the satellited-based location information to localized the tested vehicle. Also, it generates IMU data of the vehicle. The GNSS sensor collects the longitude and latitude, the angle from magnetic north, velocity, acceleration, and angular velocity of the ego vehicle. Lastly, the vision sensor (Mobileye ELD) is located on the vehicle’s front windshield. This sensor is used to collect the lane data. Based on the above data collection platform setup, demonstrations have been collected. 

\begin{figure}[t]
  \centering
  \includegraphics[width=0.8\linewidth]{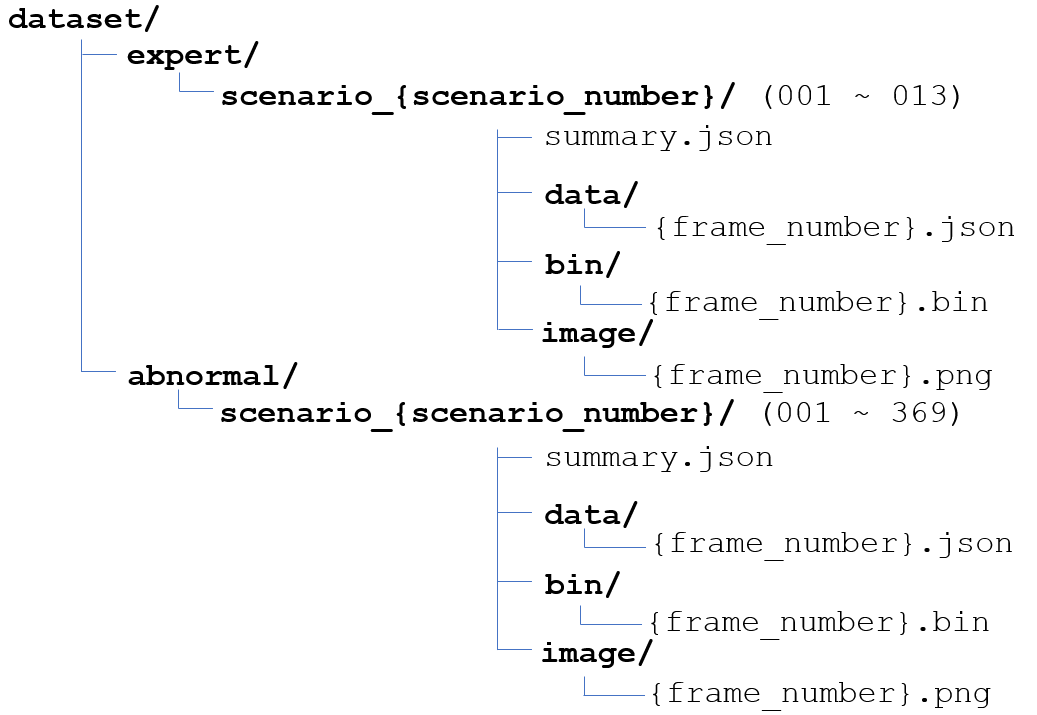} 
\caption{$\mathbf{Data~Structure.}$ The figure indicates the data structure of the provided dataset. In the expert summary file, the data collection locations are listed. The road type and the hazard type are listed in a dictionary structure in the abnormal summary file.}
\label{fig:structure}
\end{figure}

\subsection{Data Collection Method} 
We collect the proposed dataset in three different locations: highway, urban road, and Future Mobility Technology Center (FMTC). The data structure of the collected dataset is shown in Figure \ref{fig:structure}.

\begin{figure}[t]
  \centering
  \includegraphics[width=0.9\linewidth]{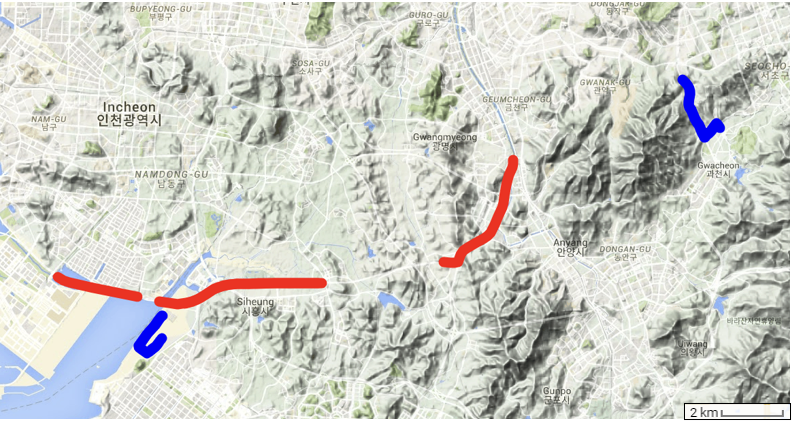} 
\caption{$\mathbf{Recording~Zone.}$ The map shows the GPS information during the collection of highway and road data. The red and blue tracks indicate where we recorded the highway data and the urban road data, respectively. As tunnels weaken the GPS signal, we excluded the tunnel data in the highway data. Also, we excluded the FMTC map because the test site is small compared to other roads.}
\label{fig:map}
\end{figure}

\begin{table}[t]
\begin{center}
\begin{tabular}{|c|c|c|c|c|}
\hline
\multicolumn{1}{|l|}{} & location & \# of exp & \# of frames & Total \\ \hline
                       & FMTC     & 5         & 18,400       &       \\ \cline{2-4}
expert                 & urban    & 4         & 12,000       & $\mathbf{42,400}$ \\ \cline{2-4}
                       & highway  & 4         & 12,000       &       \\ \hline
\end{tabular}
\caption{$\mathbf{Expert~Demonstrations.}$ The number of episodes and frames collected in each location. \# of exp indicates the number of experiments done in each location. \# of frames indicates the number of frames collected in each location. Total indicates the total number of frames collected.}
\label{tab:expert}
\end{center}
\end{table}

\begin{figure}[t]
  \centering
  \includegraphics[width=0.9\linewidth]{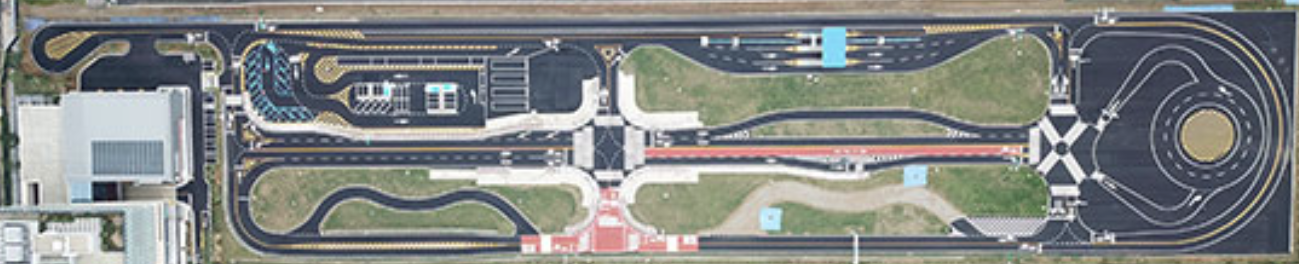} 
\caption{$\mathbf{FMTC~Snapshot.}$ The snapshot shows the autonomous driving test site in FMTC, Siheung. The track is divided into straight road, crossroad, roundabout, toll gate, overpass, and parking lot. In this paper, only the straight road, crossroad, and roundabout are used.}
\label{fig:FMTC}
\end{figure}

\subsubsection{Expert Dataset}

The expert data are collected from the areas colored in red and blue in Figure \ref{fig:map}. Figure \ref{fig:map} is a satellite image of Incheon and Seoul in Korea. Expert data comprises three significant segments; a highway, city road, and test road in FMTC. The highway data, which are marked with the red line, are collected on the highway in Gwanak (Seoul), Siheung, and Incheon. The urban road data are collected around the FMTC in Siheung, marked with the blue line. Finally, the test road data are collected on the FMTC test site, in which the experiments are artificially designed. The photo of the FMTC track is shown in Figure \ref{fig:FMTC}. The number of frames collected in each category is shown in Table \ref{tab:expert}.

\begin{table*}[t]
\centering
\begin{tabular}{|c|c|c|c|c|c|c|}
\hline
         & road type     & hazard type             & \# of exp & \# of frames & sub total & Total \\ \hline
         &               & unstable driving                & 45        & 6,360        &           &       \\ \cline{3-5}
         &               & failing lane keeping       & 49        & 6,650        &           &       \\ \cline{3-5}
         & straight road & dangerous lane changing & 50        & 1,785        & 13,300     &       \\ \cline{3-5}
abnormal &               & dangerous overtaking              & 30        & 925         &           & $\mathbf{20,355}$ \\ \cline{3-5}
         &               & near collision               & 96        & 5,620        &           &       \\ \cline{2-6}
         &               & unstable driving                & 28        & 1,380        &           &       \\ \cline{3-5}
         & crossroad     & failing lane keeping       & 8         & 530         & 7,055      &       \\ \cline{3-5}
         &               & near collision               & 180       & 5,145        &           &       \\ \hline
\end{tabular}
\caption{$\mathbf{Abnormal~Demonstrations.}$ The number of episodes and frames collected in each road and hazard type. \# of exp indicates the number of experiments done about each hazard type. \# of frames indicates the number of frames collected in each category. Subtotal and total each indicate the total number of frames collected in each road type and the total number of collected frames.}
\label{tab:negative}
\end{table*}

\subsubsection{Abnormal Dataset}
As a safety system is needed during the abnormal data collection, abnormal data are collected in the FMTC. In addition, we would like to emphasize the fact that no one is injured while collecting the abnormal driving demonstrations. The abnormal driving demonstrations are based on common accident data set (CADaS) \cite{cadas}, which provides common accident situations in Europe. To secure safety during the data collection, only less risky situations are executed.

A total of 20,455 frames of abnormal data in 369 experiments are collected during the data collection. The abnormal data is categorized in two different ways: the road type and the hazard type. The road type category includes straight road and crossroad. The hazard type category includes unstable driving, failing lane keeping, dangerous lane changing, dangerous overtaking, and near collision. Note that the hazard type categorization is not entirely distinguishable from each other since the driving situations cannot be strictly divided into separate categories. 

In the case of an abnormal situation on the straight road, 13,300 frames of data from 153 situations are collected. The abnormal situations include an ego vehicle moving straight while occupying two lanes, rapid acceleration, rapid deceleration, high angular velocity, near-collision with a stopped vehicle, near-collision while overtaking, near-collision while lane changing, near-collision with an oncoming vehicle, and near-collision with a U-turn vehicle. The exact number of frames for each situation is shown in Table \ref{tab:negative}.

In the case of abnormal situations at the crossroad, 7,055 frames of data from 216 situations are collected. The abnormal situations on the crossroad include accidents occurring when the driver ignores the traffic rules at crossroad. However, we do not collect dangerous lane changing or overtaking cases due to safety issues. The exact number of frames for each situation is shown in Table \ref{tab:negative}.

\subsection{Anomaly Detection}
The experiments about discriminating expert and abnormal demonstrations are executed to evaluate the collected dataset and provide a baseline for future researchers. As explained in Section \ref{intro}, the expert dataset and the abnormal dataset can be seen as an ID and OOD relationship. In order to discriminate the OOD data from ID data, anomaly detection algorithms are used.

For anomaly detection, the uncertainties from an MDN and the reconstruction loss from a VAE are used as the measurement of anomaly detection. For MDN, we used 10 Gaussian mixtures, where each mixture is a two-layered network. The size of each layer is 128. For VAE, the encoder and decoder are two-layered network, with the size of each layer being 128, and the encoded dimension being 32. Both algorithms used the Adam optimizer \cite{ADAM} to optimize and used weight decay of 1e-4. The batch size and the learning rate are set to 128 and 1e-3, respectively. The number of epochs trained is 100, and after the 50th epoch, the learning rate is decayed to 9e-4. The state data is composed of two parts. The first part is the linear velocity, acceleration, angular velocity, lane deviation, and the movement decision of the ego vehicle. The second part is the state data of other vehicles, composed of each vehicle's relative position and orientation from the ego vehicle, the linear velocity, the linear acceleration, and the angular velocity. The relative positions of other vehicles are each composed of the lineal and lateral distance from the ego vehicle and the relative angle between the heading of the ego vehicle and the other vehicle. 

As the state dimension varies due to the varying number of surrounding vehicles, the maximum number of the other vehicle is set to five. When the number of other vehicles is less than five, dummy objects fill up the vacant spaces. In the case of the relative positions and the angular velocities, the values of the dummy objects are set to zero. In the case of the linear velocity and the linear acceleration, the values of the dummy objects are set to corresponding values of the ego vehicle. Therefore, the dimension of the state data is 33, which is composed of 30 features from five other vehicle features (six dimensions each), and three features from the ego vehicle (linear velocity, lane deviation, and movement decision of the ego vehicle). For action data, the angular velocity and the linear acceleration of the ego vehicle are used. The dimension of the action data is two.

Two different experiments are executed to show the performance difference between single-frame state-action pairs and multi-frame sequential state-action pairs for training data. The input data for single-frame state-action pairs for both MDN and VAE are the state data. The output data of MDN are the action data. In the case of using the multi-frame sequential state-action pairs (five-frame pairs are used in experiments), the input data for both MDN and VAE is the five state data and the first four action data. The output data for MDN are the last (fifth) action data. As the number of input dimensions is increased, the network sizes of VAE are changed to 256.

\section{experiment results}
\begin{figure*}[t!]
  \centering
  \includegraphics[width=\linewidth]{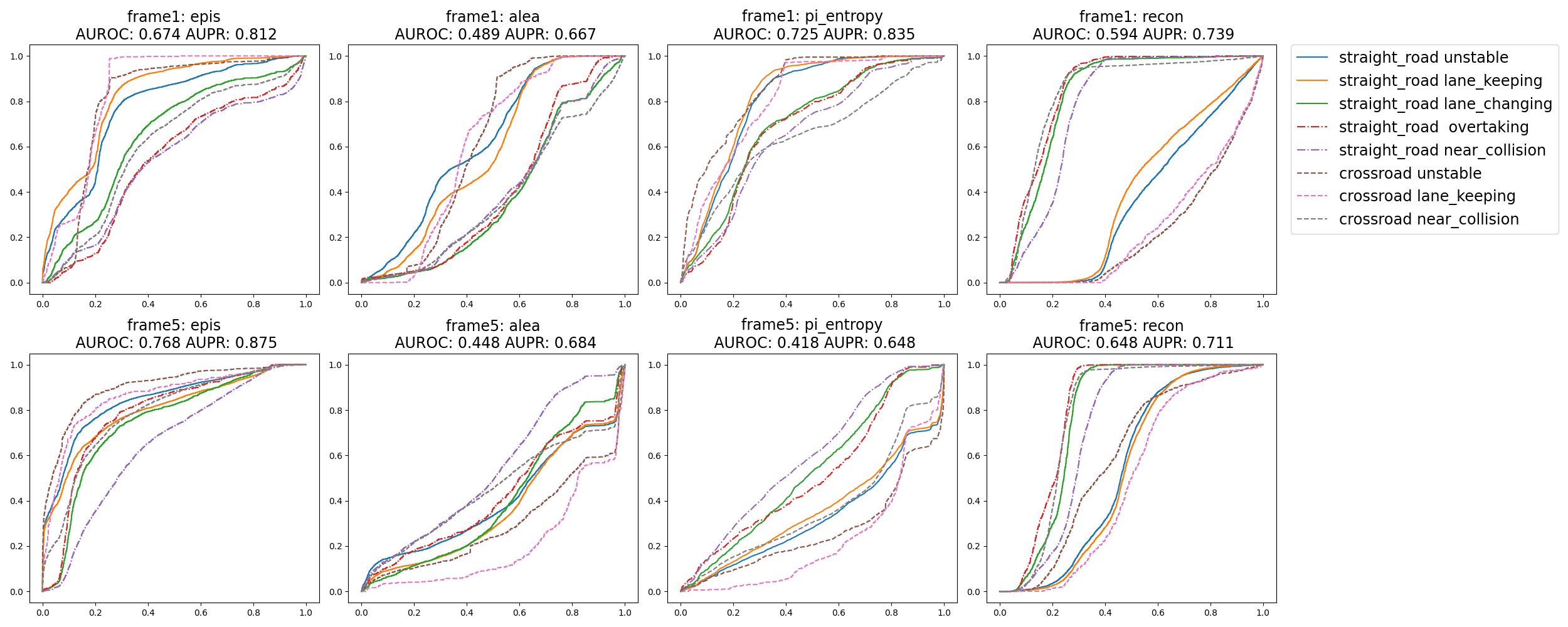}
\caption{$\mathbf{Anomaly~Detection~Results.}$ The results of using the uncertainties and reconstruction loss using a single frame and five consecutive frames for anomaly detection. Epis, alea, pi\_entropy, and recon each indicate the epistemic uncertainty, the aleatoric uncertainty, the pi-entropy, and the reconstruction loss. The lane\_keeping indicates the failing lane-keeping case. The overtaking indicates the dangerous overtaking case. The lane\_changing indicates the dangerous lane changing case.}
\label{fig:uncres}
\end{figure*}

\begin{table*}[t]
\begin{tabular}{|c|c|c|c|c|c|c|c|c|}
\hline
           & \begin{tabular}[c]{@{}c@{}}unstable\\ driving\\ (straight)\end{tabular} & \begin{tabular}[c]{@{}c@{}}failing\\ lane keeping\\ (straight)\end{tabular} & \begin{tabular}[c]{@{}c@{}}near-collision\\ (straight)\end{tabular} & \begin{tabular}[c]{@{}c@{}}unstable\\ driving\\ (cross)\end{tabular} & \begin{tabular}[c]{@{}c@{}}failing\\ lane keeping\\ (straight)\end{tabular} & \begin{tabular}[c]{@{}c@{}}dangerous\\ lane changing\\ (straight)\end{tabular} & \begin{tabular}[c]{@{}c@{}}dangerous\\ overtaking\\ (straight)\end{tabular} & \begin{tabular}[c]{@{}c@{}}near-collision\\ (cross)\end{tabular}  \\ \hline
Epistemic  & \begin{tabular}[c]{@{}c@{}}0.769/0.686\\ \textbf{0.843}/\textbf{0.818}\end{tabular}       & \begin{tabular}[c]{@{}c@{}}\textbf{0.824}/0.758\\ 0.801/\textbf{0.788}\end{tabular}           & \begin{tabular}[c]{@{}c@{}}0.533/0.406\\ 0.647/0.456\end{tabular}   & \begin{tabular}[c]{@{}c@{}}0.799/0.270\\ \textbf{0.901}/\textbf{0.681}\end{tabular}    & \begin{tabular}[c]{@{}c@{}}0.842/0.166\\ \textbf{0.860}/\textbf{0.340}\end{tabular}           & \begin{tabular}[c]{@{}c@{}}0.649/0.234\\ 0.746/0.283\end{tabular}              & \begin{tabular}[c]{@{}c@{}}0.541/0.099\\ 0.783/0.194\end{tabular}           & \begin{tabular}[c]{@{}c@{}}0.607/0.419\\ 0.788/0.646\end{tabular} \\ \hline
Aleatoric  & \begin{tabular}[c]{@{}c@{}}0.624/0.476\\ 0.402/0.414\end{tabular}       & \begin{tabular}[c]{@{}c@{}}0.577/0.446\\ 0.369/0.374\end{tabular}           & \begin{tabular}[c]{@{}c@{}}0.401/0.323\\ 0.536/0.416\end{tabular}   & \begin{tabular}[c]{@{}c@{}}0.610/0.152\\ 0.294/0.094\end{tabular}    & \begin{tabular}[c]{@{}c@{}}0.602/0.065\\ 0.205/0.032\end{tabular}           & \begin{tabular}[c]{@{}c@{}}0.379/0.135\\ 0.400/0.124\end{tabular}              & \begin{tabular}[c]{@{}c@{}}0.410/0.085\\ 0.422/0.078\end{tabular}           & \begin{tabular}[c]{@{}c@{}}0.376/0.291\\ 0.448/0.338\end{tabular} \\ \hline
Pi-entropy & \begin{tabular}[c]{@{}c@{}}0.804/0.666\\ 0.329/0.334\end{tabular}       & \begin{tabular}[c]{@{}c@{}}0.820/0.692\\ 0.357/0.355\end{tabular}           & \begin{tabular}[c]{@{}c@{}}0.655/0.499\\ 0.593/0.433\end{tabular}   & \begin{tabular}[c]{@{}c@{}}0.857/0.455\\ 0.265/0.088\end{tabular}    & \begin{tabular}[c]{@{}c@{}}0.807/0.156\\ 0.228/0.033\end{tabular}           & \begin{tabular}[c]{@{}c@{}}0.699/0.274\\ 0.530/0.159\end{tabular}              & \begin{tabular}[c]{@{}c@{}}0.685/0.147\\ 0.516/0.096\end{tabular}           & \begin{tabular}[c]{@{}c@{}}0.642/0.532\\ 0.370/0.276\end{tabular} \\ \hline
Recon loss & \begin{tabular}[c]{@{}c@{}}0.363/0.332\\ 0.548/0.399\end{tabular}       & \begin{tabular}[c]{@{}c@{}}0.402/0.357\\ 0.536/0.402\end{tabular}           & \begin{tabular}[c]{@{}c@{}}\textbf{0.778}/\textbf{0.554}\\ 0.718/0.467\end{tabular}   & \begin{tabular}[c]{@{}c@{}}0.227/0.084\\ 0.593/0.140\end{tabular}    & \begin{tabular}[c]{@{}c@{}}0.243/0.035\\ 0.492/0.049\end{tabular}           & \begin{tabular}[c]{@{}c@{}}\textbf{0.827}/\textbf{0.349}\\ 0.774/0.255\end{tabular}              & \begin{tabular}[c]{@{}c@{}}\textbf{0.850}/\textbf{0.248}\\ 0.804/0.169\end{tabular}           & \begin{tabular}[c]{@{}c@{}}\textbf{0.835}/\textbf{0.622}\\ 0.779/0.479\end{tabular} \\ \hline
\end{tabular}
\caption{$\mathbf{Anomaly~Detection~Results.}$ The numerical result of each experiment. The first line of each row indicates the AUROC (left) and AUPR (right) scores calculated in the experiments using a single frame. The second line of each row indicates the AUROC (left) and AUPR (right) scores calculated in the experiments using five consecutive frames. The bold figure indicates the best AUROC and AUPR scores for each situation.}
\label{tab:uncnum}
\end{table*}

The comparative visual results of epistemic uncertainty, aleatoric uncertainty, pi-entropy, and reconstruction loss using one frame and five consecutive frames are shown in Figure \ref{fig:uncres}. The numerical results of the experiments are shown in Table \ref{tab:uncnum}. For the evaluation methods, the AUROC and the AUPR score are used.

From the results shown in Table \ref{tab:uncnum}, it can be inferred that the aleatoric uncertainty cannot discriminate the abnormal situations. The AUROC and the AUPR scores of aleatoric uncertainty are less than 0.65, which illustrates that aleatoric uncertainty is not a suitable anomaly detection method. In addition, as the number of frames increases, all the scores drop below 0.5, which is worse than random selection.

In the case of the pi-entropy, it shows the best results among the algorithms at discriminating the unstable driving cases collected in both crossroad and straight road with single-frame pairs, as shown in Table \ref{tab:uncnum}. However, the discrimination quality sharply drops when using five consecutive frames. 

In the case of the epistemic uncertainty, it shows the best results among the algorithms at the discriminating failing lane-keeping cases collected in both crossroad and straight road when using a single frame, as shown in Table \ref{tab:uncnum}. The discrimination ability is reinforced when using five consecutive frames, excelling at the unstable driving cases and failing lane-keeping cases, which are the best results among the entire experiments.

In the case of the reconstruction loss, it shows the best results among the algorithms at the discriminating near-collision cases collected in both crossroad and straight road, and the dangerous lane changing and overtaking cases in a straight road when using a single frame, as shown in Table \ref{tab:uncnum}. These results are the best results among the entire experiments in the above four cases. Unlike epistemic uncertainty, the results of the reconstruction loss deteriorate when the number of frames increases to five.

An interesting part of the result is that the reconstruction loss and epistemic uncertainty excel at different cases. In the case of MDN, the network excels at discriminating unstable driving and failing lane-keeping cases. When the epistemic uncertainty is high, the action of the ego vehicle has high variance. Hence, it can be inferred that the ego vehicle is under unstable driving or failing lane-keeping. On the other hand, the reconstruction error excels at discriminating near-collision, dangerous lane changing, and overtaking cases. The reconstruction error increases when the surroundings of the ego vehicle are different from the data used to train the network. Therefore, it can be inferred that near-collision, dangerous lane changing, and overtaking are mainly affected by the surrounding of the ego vehicle. In conclusion, the epistemic uncertainty is useful at detecting abnormal behaviors of the ego vehicles while the reconstruction loss is useful for detecting abnormal surroundings. 

When using five consecutive frames, the performance of the epistemic uncertainty increases, as the information from the past is also given to the network. This increased information allows the epistemic uncertainty to consider the sequential results of the action during prediction. This allows the epistemic uncertainty to better discriminate abnormal cases. On the other hand, the performance of the reconstruction loss decreases when using five consecutive frames. 
As the VAE input has changed into consecutive frames, the results of some of the data that has been viewed abnormal are changed. This change of view leads the reconstruction error to lessen, causing deterioration in AUROC and AUPR scores.

From the experimental results, it is desirable to use the epistemic uncertainty with multiple consecutive frames and the reconstruction loss with a single current frame for anomaly detection, as they cover all the abnormal cases. From the results, it can be inferred that unstable driving and failing lane-keeping cases can be regarded as problems with the ego-driver, as the epistemic uncertainty better captures these cases. On the other hand, near-collision, dangerous overtaking, and lane changing cases can be regarded as problems related to surrounding drivers, as the reconstruction loss better captures these cases.

\section{Conclusion}
This paper presents a novel dataset called the \textit{R3 Driving Dataset}, which contains normal and abnormal driving demonstrations. The number of normal and abnormal demonstrations is 42,400 and 20,355, respectively. The data collection platform is also proposed, along with baseline methods for anomaly detection of the proposed dataset. The MDN and VAE are used for anomaly detection, and the evaluation metrics of the algorithm are the AUROC and AUPR. From the results, we can infer that near-collision, dangerous lane changing, and overtaking cases can be detected effectively using the reconstruction loss, while unstable driving and failing lane-keeping cases can be detected more effectively using the epistemic uncertainty.

\bibliographystyle{IEEEtran}
\bibliography{myBIB}

\addtolength{\textheight}{-12cm}   



\end{document}